\def\year{2019}\relax
\documentclass[letterpaper]{article} 
\usepackage{aaai19}  
\usepackage{times}  
\usepackage{helvet}  
\usepackage{courier}  
\usepackage{url}  
\usepackage{graphicx}  
\usepackage{times}
\usepackage{xcolor}
\usepackage{soul}
\usepackage[utf8]{inputenc}
\usepackage{amssymb}
\usepackage{graphicx}
\usepackage{stmaryrd}
\usepackage{booktabs}
\usepackage{array}  
\usepackage{amsmath}
\usepackage{algorithm}
\usepackage{algorithmic}

\newcounter{ALC@tempcntr}
\newcommand{\citet}[1]{\citeauthor{#1}~\shortcite{#1}}
\newcommand{\citep}{\cite}
\newcolumntype{x}[1]{>{\centering\arraybackslash\hspace{0pt}}p{#1}}

\title{A Sequential Set Generation Method for Predicting Set-Valued Outputs } 

\author{
Tian Gao, 
Jie Chen, 
Vijil Chenthamarakshan, 
Michael Witbrock\\ 
IBM Research\\
Thomas J. Watson Research Center, Yorktown Heights, NY\\
\{tgao, chenjie, ecvijil, witbroc\}@us.ibm.com
}

\begin{document}

\maketitle

\begin{abstract}
Consider a general machine learning setting where the output is a set of labels or sequences. This output set is unordered and its size varies with the input. Whereas multi-label classification methods seem a natural first resort, they are not readily applicable to set-valued outputs because of the growth rate of the output space; and because conventional sequence generation doesn't reflect sets' order-free nature. In this paper, we propose a unified framework---sequential set generation (SSG)---that can handle output sets of labels and sequences. SSG is a meta-algorithm that leverages any probabilistic learning method for label or sequence prediction, but employs a proper regularization such that a new label or sequence is generated repeatedly until the full set is produced. Though SSG is sequential in nature, it does not penalize the ordering of the appearance of the set elements and can be applied to a variety of set output problems, such as a set of classification labels or sequences. We perform experiments with both benchmark and synthetic data sets and demonstrate SSG's strong performance over baseline methods. 
\end{abstract}

\section{Introduction}
\label{intro}

Recent advances in machine learning, particularly deep learning models and training algorithms, have resulted in significant breakthroughs in a variety of AI areas, including computer vision, natural language processing, and speech recognition. Most of these applications have been formulated as classification problems: a label is predicted for a given input. The output label could be the category of an image, the word uttered in an audio signal, or the topic of a news paragraph. For sequence generation problems, an ordered list of tokens is generated sequentially, with the output of each token being essentially a label prediction. In this paper, we pursue the capability to predict sets, the size of which may vary, and for which the order of the elements is irrelevant. We call this problem \emph{set prediction}. The challenge lies in the fact that the output space, or the universe of set elements, may be enormously large or even infinite, especially for sets of sequences. Thus, treating the general problem as multi-label classification is inefficient or effectively impossible. Examples of set prediction problems include learning to enumerate relevant rules and possible bindings of a logic-based inference system, producing all descriptions of a picture, and generating relevant images for a given query. 

A major goal of our lab is to work toward unifying the capabilities of deep learning approaches with the AI capabilities supported by symbolic computation, and a major thread of such work concerns logical inference, including mathematical theorem proving. In theorem proving applications \cite{irving2016deepmath}, one needs to produce sets of complex structures representing a search state and its possible extension, and then reduction, as a solution is constructed. For example, one needs to select a set of mathematical statements relevant to finding solutions for a given conjecture, say $A(x,{\tt Volume6})$, such as $\{y\mapsto {\tt Volume6}\} \slash \{A(x,y) \leftarrow B(x) \wedge C(y,x) \wedge D(y)$,  $A(x,y) \leftarrow F(y) \wedge E(x,y)\}$. One also needs to find, and then apply, a set of bindings that satisfy at least one of the possible solution paths, such as $x\mapsto\{5,8,1\}$ supposing that $F({\tt Volume6})$ and $D({\tt Volume6})$ hold, and so do $E(1,{\tt Volume6})$, $B(5)$, $B(8)$, $C({\tt Volume6},5)$ and $C({\tt Volume6},8)$. Note that in both these cases ---finding relevant conjectures, and finding bindings that satisfy those conjectures--- what is being manipulated is a set of complex sequences representing logical formulas.

While it is conceivable for an algorithm to be trained to produce a sequence representing the relevant output set, doing so often requires the introduction of some artificial order over the elements, which is quite unnatural. Moreover, the complexity of choosing a particular "good" list order may be prohibitive, and finding this "best" order during inference may be simultaneously challenging and pointless. Recent work has shown that choosing such a "right" order is crucial for prediction performance \citep{order_matters}. 

In this work, we aim at predicting an output set (of symbols or sequences) that has bounded (but varying) size and is order-free. We propose a meta-algorithm, called Sequential Set Generation (SSG), that predicts output elements one by one until the full set is produced. SSG handles sets of labels as in the standard classification setting, as well as sets of sequences needed for rule induction, inference, or image generation. We demonstrate these two capabilities with synthetic data sets and show the empirical success of the proposed algorithm.
%
%
%


\medskip

\section{Related Work}

There are two main areas of work related to the set-valued output problem. The first is Sequence-to-Sequence models,  which have found widespread application in areas including machine translation~\cite{DBLP:journals/corr/BahdanauCB14:NMT,DBLP:journals/corr/ChoMGBSB14}, image captioning~\cite{DBLP:journals/corr/VinyalsTBE14}, and speech recognition~\cite{hintonspeech}. In these applications,  explicit orderings of input and output sequences are assumed. However, the choice of a particular ordering affects the accuracy of the algorithm. For example, \citet{DBLP:journals/corr/SutskeverVL14} report a 5 BLEU point improvement in translation from English  to French, if the order of each English sentence is reversed. Moreover, \citet{order_matters} conduct extensive experiments and demonstrate that the input/output order significantly affects performance on a variety of learning tasks, including language modeling and parsing. They also suggest ways to handle set inputs (using an attention mechanism and memory) and set outputs (searching over possible output orderings), which can quickly become intractable. 

Another related area comprises the multi-label \cite{tsoumakas2009mining,zhang2014review}, multi-task \cite{xue2007multi,argyriou2007multi,argyriou2008convex}, and structured prediction \cite{taskar2005learning} problems. Each of these problems produces multiple outputs, usually in the form of classification results. They can leverage information from other labels and share information to improve the learning of all outputs jointly, and have been widely used in many machine learning applications. While these learning methods perform very well in many applications, they have to explicitly model each output in large scale classification problems, which quickly becomes infeasible. In this work, we propose an alternative formulation that makes the problem of set prediction tractable. More importantly, our formulation is very general, not limited to classification, and can handle multiple forms of sets, including sets of sequences. 

Recently learning methods for set-valued input problems have also been investigated \cite{zaheer2017deep}, showing that there is increasing interest in this broadly-applicable class of problems.

\section{Problem Statement}

Let $\mathcal{R}^d$ be the input space and $S$ be the label space, which could possibly be countably infinite. Given $N$ data samples $X_i \in \mathcal{R}^d$, $i = 1, \ldots, N$ and corresponding outputs $Y_i \in P(S)$, where $P(S)$ denotes the power set of $S$, the objective is to learn a function $f: \mathcal{R}^d \to P(S)$ that (approximately) obeys the constraints inherent in the given data $f(X_i)=Y_i$. We assume that every output set $Y_i$ is finite. Here is an example:

\textbf{Example}: Let $f(X)=\{ \text{integer } Y : \, Y > X \text{ and } Y \le 10\}$. Given training samples $X_1=1.01$, $Y_1=\{2,3,4,5,6,7,8,9,10\}$; $X_2=5$, $Y_2=\{6,7,8,9,10\}$; and $X_3=9.5$, $Y_3=\{10\}$, predict the output $Y_4$ when $X_4=8.7$.




This simple example can be extended to many real-life applications (e.g., semantic matching, graph traversal, and question answering), where multiple outputs are required to fully answer a question. 

\section{Base Framework: Sequential Set Generation } 
To handle the variable sizes of the output sets, we split each output $Y_j$ into individual elements and reformulate the training data as $D = \{(\hat{X}_i, \hat{Y}_i) \mid i=1, \ldots, \sum_j|Y_j|\}$, where each $\hat{X}_i$ is an original $X_j$ and $\hat{Y}_i$ is an element of $Y_j$. For testing, the trained classifier should produce the entire set $Y_k$ given a test sample $X_k$.

\begin{figure}[t]
\begin{center}
\centerline{\includegraphics[width=1.0\columnwidth, scale=5]{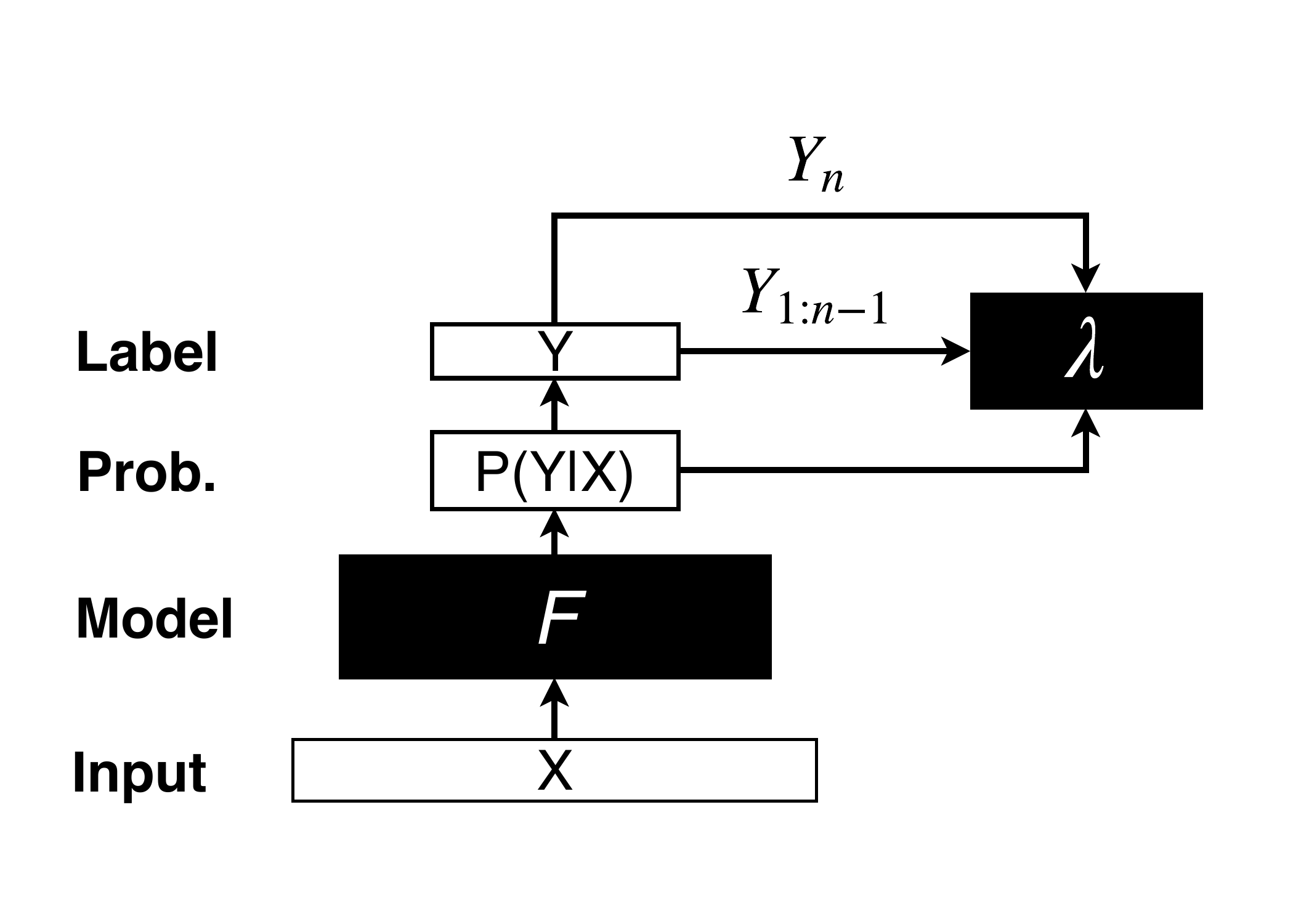}}
\caption{System Overview for SSG. Given an input $X$, SSG uses the trained probability to learn $\lambda$ via a optimizer, which sequentially generates one element of output $Y_n$ at a time, given previous outputs $Y_{1:{n-1}}$, until generating the set of all output $Y$. }
\label{fig:ssg}
\end{center}
\end{figure}

If one directly fits a model between the $\hat{X}_i$'s and $\hat{Y}_i$'s, by using, e.g., logistic regression or neural networks, the loss for the same $\hat{X}_i=X_j$ should be similar between the different $\hat{Y}_i \in Y_j$, indicating an equal probability for obtaining one of the correct class labels. These models, however, produce at most one label (subject to any tie breaking mechanism) but not the entire set. Rather than developing a new model for our problem, we propose a general framework, called Sequential Set Generation (SSG), that produces a set of labels through leveraging any existing classification models with an additional regularization. The overview of the system is shown in Figure ~\ref{fig:ssg}.

The proposed framework is suitable for any machine learning classifier and can deal with many different set prediction problems. The framework is versatile, and generalizes beyond standard label predictions to, e.g., sequence predictions, where each output $\hat{Y}_i$ (an element of the output set $Y_j$), is by itself a sequence. 
We will discuss the applications of SSG  and its generalization to sequence learning. 



The algorithm proceeds as follows. 
SSG produces set elements sequentially. At each step, we want to find the most plausible answer that has not appeared before, for which we use a memory $\mathbf{Z}$ to keep track of. Hence, the predictive output is computed as: 
%
\begin{alignat*}{2}
Y^* = \arg\min_{\hat{Y}} \quad & f_{\theta}(\hat{X},\hat{Y}) \\
  \text{s.t.} \quad &  \begin{aligned}[t]
     \hat{Y} & \not\in {\bf Z},\\
  \end{aligned}
\end{alignat*}
where $\theta$ consists of the learned parameters of a model $f$, and $\mathbf{Z}$ is the set of answers produced so far. To ease computation, we move the constraint to the objective function through Lagrange relaxation:
\begin{equation}
\label{eq:obj}
Y^* = \arg\min_{\hat{Y}}  \quad f_{\theta}(\hat{X},\hat{Y}) + \lambda I_{\hat{X}}(\hat{Y},{\bf Z}),
\end{equation}
where $\lambda$ is the coefficient for the memory penalty, and $I$ is an indicator function that penalizes a potential label of $\hat{X}$ that has already appeared in the memory ${\bf Z}$. One can use the Hamming loss, for example, to compute $I$: $I_{\hat{X}}(\hat{Y},{\bf Z}) = \sum_i I_{\hat{X}}(\hat{Y} = {\bf Z}_i)$.

 In essence, SSG utilizes the memory ${\bf Z}$ to store existing outputs and repeatedly generates plausible answers to form the output set, until a new answer repeats itself. SSG incorporates the memory penalty term to realize such a sequential process.
 


\section{ Training and Test for SSG}
In what follows, we first consider how SSG works in testing and then state the method for training.

\textbf{During Testing}: Given a query sample $\hat{X}$ and a set ${\bf Z}$ (which can be either empty or not), Equation~\eqref{eq:obj} produces the next most plausible label.  We repeatedly use~\eqref{eq:obj} until a stopping criterion is reached. To ensure all the correct output labels are produced, we use the following criterion: if $Y^*$  in~\eqref{eq:obj} exists in ${\bf Z}$, SSG terminates and outputs all the elements in ${\bf Z}$. Otherwise, SSG stores $Y^*$ into ${\bf Z}$ and compute another $Y^*$. It repeats the procedure to generate correct labels while ensuring the incorrect answers are not produced. In the end, the stored memory ${\bf Z}$ should contain the entire output set. This testing procedure is summarized in Algorithm~\ref{alg:test}. Note that in order to generate the first element of the set, we use the first term of~\eqref{eq:obj}. 

\begin{algorithm}[tb]
   \caption{SSG Algorithm Testing Procedure}
   \label{alg:test}
\begin{algorithmic}
   \STATE {\bfseries Input:} Testing data $\hat{X}$, parameters $\theta$ and $\lambda$ 
   \STATE ${\bf Z} \leftarrow \emptyset; Ans \leftarrow \emptyset;$
   \STATE $N \leftarrow$ number of testing samples $\hat{X}$
   \FOR{$i=1$ {\bfseries to} $N$}
   \STATE {$Y^* \leftarrow \arg\min_{\hat{Y}} f_{\theta}(\hat{X}_i,\hat{Y})$}

   \WHILE {$Y^*$ is not in ${\bf Z}$ }
   \STATE {${\bf Z} \leftarrow {\bf Z} \cup Y^*$}
   \STATE {$Y^* \leftarrow$ Compute Equation~\eqref{eq:obj}}
   \ENDWHILE
      \STATE {$Ans\{i\} \leftarrow {\bf Z}$}
   \ENDFOR
	 \STATE {\bfseries Return:} {$Ans$}
\end{algorithmic}
\end{algorithm}

This formulation can also answer questions such as ``what else would be a good class label given data and existing labels.'' 

\textbf{During Training}: To facilitate the application of different machine learning models, we  would like a general training procedure that is widely applicable to different loss functions. We have the following training objective: 
\begin{equation}
\label{eq:train_overall}
\theta^*, \lambda^* = \arg\min_{\theta,\lambda}  \quad \mathcal{L}(\hat{X},\hat{Y};\theta) + g(\lambda),
\end{equation}
where $\mathcal{L}(\hat{X},\hat{Y}; \theta)$ denotes the loss function of a machine learning model, given training data $\hat{X}=\{\hat{X}_i\}$ and $\hat{Y}=\{\hat{Y}_i\}$, and $g$ is a loss that corresponds to the memory penalty in \eqref{eq:obj}, which we will elaborate. The function $\mathcal{L}$ may be any loss (e.g., negative log likelihood) that is associated with the predictive model $f_{\theta}$.

We observe that the training of the two parameters in \eqref{eq:train_overall} can be separated, as the parameter $\theta$ for the model $f$ and the memory penalty parameter $\lambda$ resides on different terms. Hence, we first train the first term, equivalent to training any classifier using their specialized procedures (e.g., random forests, SVM, or neural networks). 

Then, we compute the memory penalty coefficient $\lambda$ from $g(\lambda)$. We would like the memory term to penalize wrong predictions while promoting correct ones. While there exist many  choices satisfying this requirement, we use the max-margin principle; i.e., maximizing the gap between the stored labels and other correct labels, as well as those between the stored labels and incorrect labels. We propose the following training objective for robust estimation of $\lambda$:
\begin{alignat*}{2}
 \lambda^* = \arg\min_{\lambda}  \quad & \sum_i^N [P(\hat{Y}_i|\hat{X}_i) - \hat{P}_i - \lambda]^2 \\
  \text{s.t.} \quad &  \begin{aligned}[t]
    P(\hat{Y}_i|\hat{X}_i) - \lambda & \ge L^{-}_{max, \hat{X}_i}, \,\,\,\forall i\\
    P(\hat{Y}_i|\hat{X}_i) - \lambda & \le L^{+}_{min, \hat{X}_i}, \,\,\,\forall i\\
  \end{aligned}
\end{alignat*}
where $P(\hat{Y}_i|\hat{X}_i)$ denotes the posterior probability resulting from the trained model, $L^{-}_{max, \hat{X}_i}$ (resp. $L^{+}_{min, \hat{X}_i}$) is the maximal (resp. minimal) posterior probability of the set of negative (resp. positive) labels for  $\hat{X}_i$, and $\hat{P}_i$ is the average between them; i.e., $\hat{P}_i = (L^{-}_{max, \hat{X}_i} + L^{+}_{min, \hat{X}_i})/2$.


The above equation can be solved by using Lagrangian relaxation, leading to: 

\begin{multline}
\label{eq:train_ll}
\lambda^* = \arg\min_{\lambda}  \sum_i^N ||P(\hat{Y}_i|\hat{X}_i) - \hat{P}_i - \lambda||_2^2 \\
 - \lambda_+ [P(\hat{Y}_i|\hat{X}_i) - \lambda - L^{-}_{max, \hat{X}_i}] \\
 + \lambda_- [ P(\hat{Y}_i|\hat{X}_i) - \lambda  - L^{+}_{min, \hat{X}_i}]
\end{multline}
where $\lambda_+$ and $\lambda_-$ are the Lagrangian multipliers of the two constraints. They can be set to large values to ensure satisfaction of constraints. 
%
%
%
%

The analytical solution of Equation~\ref{eq:train_ll}  is that $\lambda^*$ is either on the boundary
\begin{gather*}
\min_i\left\{P(\hat{Y}_i|\hat{X}_i)-L^{-}_{max, \hat{X}_i}\right\},\\
\max_i\left\{P(\hat{Y}_i|\hat{X}_i)-L^{+}_{min, \hat{X}_i}\right\},
\end{gather*}
or is equal to the unconstrained minimizer
\[
\frac {\sum_i^N  P(\hat{Y}_i|\hat{X}_i) - \hat{P}_i}{N}
\]
if it is feasible, whichever achieves a lower objective value. See Algorithm~\ref{alg:train}.

\begin{algorithm}[tb]
   \caption{SSG Algorithm Training Procedure}
   \label{alg:train}
\begin{algorithmic}
   \STATE {\bfseries Input:} Training data $\hat{X}$, training labels $\hat{Y}$
   \STATE $\theta^* \leftarrow \arg\min_{\theta} \mathcal{L}(\hat{X},\hat{Y};\theta);$
   \FOR{each unique $\hat{X}_i$ in $\hat{X}$}
  \STATE $Y^+ \leftarrow \hat{Y}[\hat{X}_i]$
  \STATE Compute $L^{+}_{min, \hat{X}_i}$ using $Y^+$ 
  \STATE $Y^- \leftarrow \hat{Y}\backslash Y^+$
  \STATE Compute $L^{-}_{max, \hat{X}_i}$ using $Y^-$ 
  \STATE $ \hat{P}_i \leftarrow (L^{-}_{max, \hat{X}_i} + L^{+}_{min, \hat{X}_i})/2$
  \STATE Compute $P(\hat{Y}_i|\hat{X}_i) - \hat{P}_i$ 
   \ENDFOR
   \STATE Choose $\lambda^*$ from unconstrained minimizer or boundary
	 \STATE {\bfseries Return:} {$\theta^*, \lambda^*$}
\end{algorithmic}
\end{algorithm}

After training the model parameter $\theta$,  we find the positive label set $Y^+$ and negative label set $Y^-$ for each training data $\hat{X}_i$. We compute the posterior probabilities for each element of $Y^+$ and $Y^-$. To follow the max-margin principle, we compute the loss gap for each $\hat{X}_i$ and set the feasible region to be the intersection of all gaps. Finally, $\lambda$ is chosen among the boundary of the feasible region and the unconstrained minimizer, whichever is feasible and achieves minimum.

In testing, for each $x_i$, we first compute the first term of the classification loss,  obtaining one label $y_{ij}$. We then penalize the loss of $y_{ij}$ by computing Equation~\eqref{eq:train_overall} and attempt  to obtain another answer $y_{ik}$, if $y_{ik}$ has not appeared in the answers. Repeated application of Equation~\eqref{eq:train_overall} until replication in the answers gives the full set of elements. 

\subsection{Stopping Criterion}

 The while-loop in Algorithm~\ref{alg:test} effectively states that if the computed label is not in the memory $\bf{Z}$, then one should continue producing more. This hard criterion may encounter problems in practice with noisy data. Here, we propose a more robust stopping criterion, which does not affect the behavior of Algorithm~\ref{alg:test} under ideal conditions. 

In addition to the memory $\bf{Z}$, we maintain a counter $C_i$ indicating the number of times a label $y_i$ is produced. Hence, the predictive function \eqref{eq:obj} now becomes:
\begin{multline}
 \label{eq:obj2}
 Y^* = \arg\min_{\hat{Y}}  \quad f_{\theta}(\hat{X},\hat{Y}) + \sum_i C_i \cdot \lambda I_{\hat{X}}(\hat{Y}={\bf Z}_i).
 \end{multline}
 Let $C$ be the vector of the same dimension as ${\bf Z}$. If $C$ is a vector of all ones, Equation~\eqref{eq:obj2} is equivalent to \eqref{eq:obj}. When the elements of $C$ are greater than $1$, the new criterion does not immediately terminate the loop; rather, the loop continues until a certain percentage of the labels have appeared in the memory more than once. In other words, if $\sum_i C_i \geq (1 + \rho) |{\bf Z}|$, where $\rho$ is a predefined value with $ 0 \leq \rho < 1$, Algorithm~\ref{alg:test} stops. In a well-trained system, the new stopping criterion will always yield at least one of the true positive labels with a lower objective value than the negative labels. With a judicious choice of $\rho$, the system becomes more robust against noise.

\section{Sequential Set Generation for Sequences}
\label{sec:seq}
The preceding section proposes a method when the output is a set, such as a set of class labels. In many applications, especially natural language problems, however, the elements of the output set are sequences (e.g., sentences), which by themselves are ordered lists comprising sub-elements (e.g., words). In this case, the SSG algorithm proposed so far cannot directly handle sequences, because sequence generation methods (e.g., sequence-to-sequence models~\cite{DBLP:journals/corr/BahdanauCB14:NMT,DBLP:journals/corr/ChoMGBSB14}) are iterative and there is no loss associated with the entire sequence.  Penalizing the entire sequence with a single $\lambda$ is not sensible.

\begin{figure}[t]
\begin{center}
\centerline{\includegraphics[width=1.1\columnwidth, scale=1]{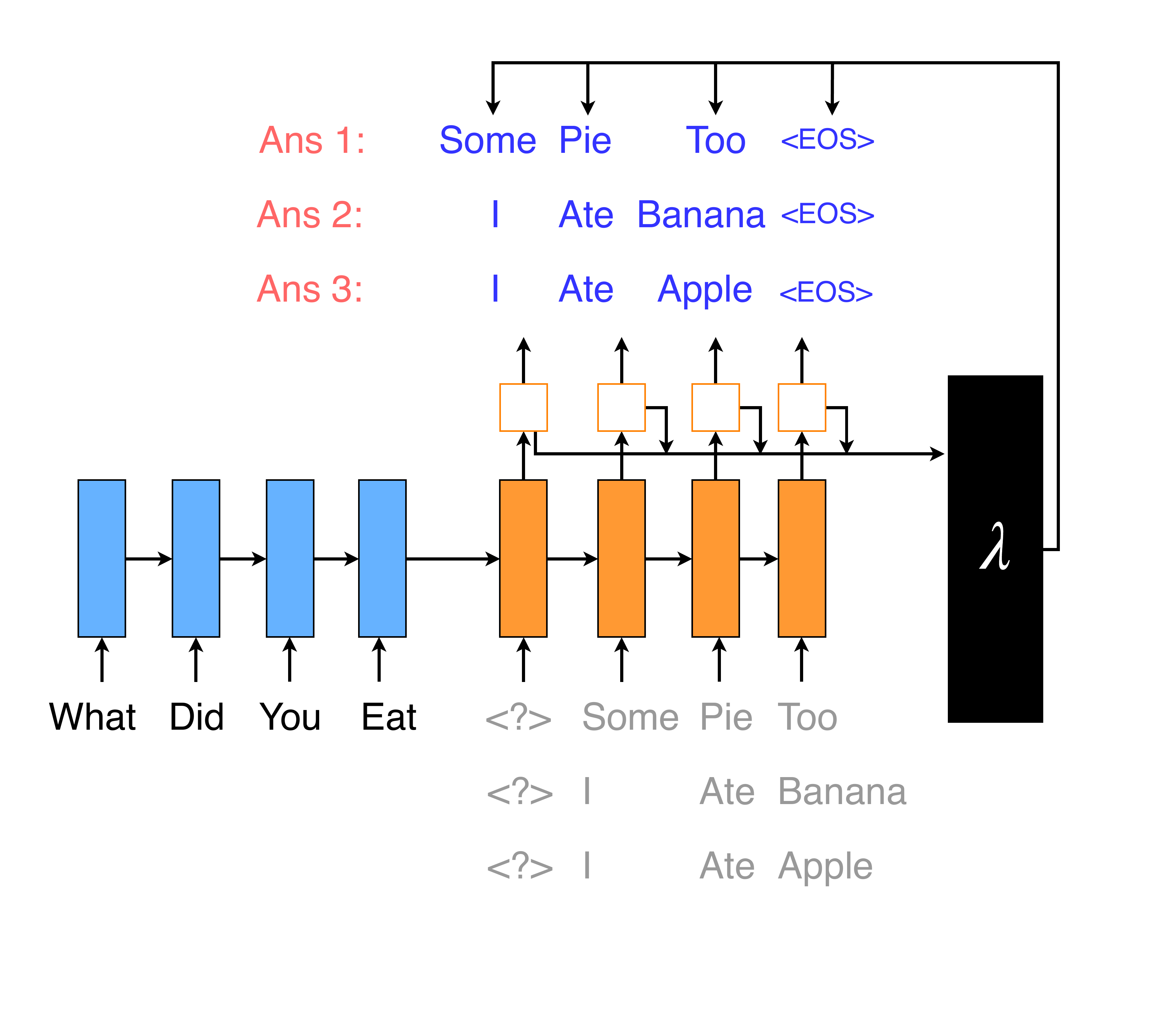}}
\caption{System Overview for SSG-S. Compared with SSG, SSG-S specifically uses a model that can model sequence inputs (such as encoder-decoder networks) to learn the relationships and use a optimizer or classifier to learn many different $\lambda$'s. }
\label{fig:ssgs}
\end{center}
\end{figure} 

We would like to extend SSG to outputs that are sets of sequences. The proposed extension is called SSG-S, and its overall architecture is shown in Figure~\ref{fig:ssgs}. The key idea is to penalize each sub-element, instead of the entire sequence, from repeating itself at each location of the output. To achieve so, we need a separate $\lambda_i$ for each output location. Let $\vec{Y} $ be one sequence output and let $\vec{Y}_i$ be an element within the sequence. Given previously generated elements $\vec{Y}_{1:i-1}$, we generate the next element $\vec{Y}_i$ as
\begin{multline}
\label{eq:objs}
\vec{Y}^*_i = \arg\min_{\vec{Y}_i}  \quad f_{\theta}(\hat{X},\vec{Y}_i|\vec{Y}_{1:i-1}) + \lambda_i I(\vec{Y}_i,{\bf Z}_i),
\end{multline}
where ${\bf Z}_i$ contains all the $i$-th elements of the stored outputs. The first term of~\eqref{eq:objs} is a typical sequence-to-sequence (seq2seq) model, which must be conditioned on the past outputs $\vec{Y}_{1:i-1}$. At each step, it produces a new element given the already produced partial sequence. The second term penalizes the elements that have appeared in the stored output. For each location of the sequence, the penalty is different. 

Similar to the preceding section, the model parameter $\theta$ and the penalty parameters $\Lambda=\{\lambda_i\}$ are trained by using the objective
\begin{equation*}
\theta^*, \Lambda^* = \arg\min_{\theta,\Lambda}  \quad \mathcal{L}(\hat{X},\hat{Y};\theta) + g(\Lambda),
\end{equation*}
where $(\hat{X},\hat{Y})$ denotes the training data and $\mathcal{L}$ is any loss in a seq2seq model that comes with the predictive function $f_{\theta}$ in~\eqref{eq:objs}. The training of $\theta$ is standard. The second term $g(\Lambda)$ is used to train the penalty parameters $\Lambda=\{\lambda_i\}$. For each location $i$ in the output sequence, $\lambda_i$ is trained by using, again, the max-margin principle through
\begin{alignat*}{2}
 \lambda_i^* = \arg\min_{\lambda_i}  \quad & \sum_j^N [P(\vec{Y}_{j,i}|\hat{X}_j,\vec{Y}_{j,1:i-1}) - \hat{P}_{j,i} - \lambda_i]^2 \\
  \text{s.t.} \quad &  \begin{aligned}[t]
    P(\vec{Y}_{j,i}|\hat{X}_j,\vec{Y}_{j,1:i-1}) - \lambda_i & \ge L^{-}_{max, \hat{X}_j, i}, \,\,\,\forall j\\
    P(\vec{Y}_{j,i}|\hat{X}_j,\vec{Y}_{j,1:i-1}) - \lambda_i & \le L^{+}_{min, \hat{X}_j, i}, \,\,\,\forall j.\\
  \end{aligned}
\end{alignat*}
The solution is similar to that in the preceding section, for each $i$.

The training and testing algorithms are shown in Algorithms~\ref{alg:trains} and ~\ref{alg:tests}, respectively. The training of SSG-G is similar to SSG, and the only difference is that the $\lambda$'s are computed for each token level in a sequence, resulting in a total of $\max |\vec{Y}|$ number of $\lambda$. The notation $\max |\vec{Y}|$ represents the maximal allowable sequence length in any of the outputs. 

\begin{algorithm}[tb]
   \caption{SSG-S Algorithm Training Procedure}
   \label{alg:trains}
\begin{algorithmic}
   \STATE {\bfseries Input:} Training data $\hat{X}$, training sequences $\hat{Y}$
   \STATE $\theta^* \leftarrow \arg\min_{\theta} \mathcal{L}(\hat{X},\hat{Y};\theta);$
   \FOR{each unique $\hat{X}_i$ in $\hat{X}$}
     \STATE $Y^+ \leftarrow \hat{Y}[\hat{X}_i]$
  \STATE $Y^- \leftarrow \hat{Y}\backslash Y^+$	  	
  \FOR{ $j = 1$ to $ \max |\vec{Y}|$ }  
  
  \STATE Compute $L^{+}_{min, \hat{X}_i, j}$ using $Y^+$
  \STATE Compute $L^{-}_{max, \hat{X}_i, j}$ using $Y^-$
  \STATE $ \hat{P}_{i,j} \leftarrow (L^{-}_{max, \hat{X}_i, j} + L^{+}_{min, \hat{X}_i, j})/2$
  \STATE Compute $P(\hat{Y}_{i,j}|\hat{X}_i,\vec{Y}_{i,1:j-1}) - \hat{P}_{i,j}$

  \ENDFOR
   \ENDFOR
  \FOR{ $j = 1$ to $ \max |\vec{Y}|$ }  
   \STATE Choose $\lambda_j^*$ from unconstrained minimizer or boundary
  \ENDFOR

	 \STATE {\bfseries Return:} {$\theta^*, \{\lambda_j^*\}$}
\end{algorithmic}
\end{algorithm}

\begin{algorithm}[h]
   \caption{SSG-S Algorithm Testing Procedure}
   \label{alg:tests}
\begin{algorithmic}
   \STATE {\bfseries Input:} Testing data $\hat{X}$, parameter $\theta$ and  $\Lambda=\{\lambda_j\}$ 
   \STATE $ Ans \leftarrow \emptyset;$
   \STATE $N \leftarrow$ number of testing samples $\hat{X}$
   \FOR{$i=1$ {\bfseries to} $N$}

      \STATE {${\bf A} \leftarrow \emptyset$;}
        \FOR{$j=1$ {\bfseries to} $\max|\vec{Y}|$}
           \STATE {${\bf Z} \leftarrow \emptyset$;}
	\FOR {each element $A_k$ in $A$}
 
   \STATE $\vec{Y}_j = \emptyset$ 
      \WHILE {$\vec{Y}_j$ is not in ${\bf Z}$ }
   \STATE {$\vec{Y}_j \leftarrow$ Compute Equation~\eqref{eq:objs}}
   \ENDWHILE
  
   \STATE {$A_k$ $\leftarrow$ Append each element $\vec{Y}$ to $A_k$}
   \ENDFOR
      \ENDFOR
      \STATE {$Ans\{i\} \leftarrow {\bf Z}$}
   \ENDFOR
	 \STATE {\bfseries Return:} {$Ans$}
\end{algorithmic}
\end{algorithm}

SSG-S has noticeable differences in testing from SSG. Specifically, SSG-S does not generate one sequence in its entirety before generating the next one. On the contrary, it generates all possible answers for each position in a sequence. This approach allows efficient data structures if desired, such as a Trie-tree, to keep track of all the sequences in the set, although it is also capable of sequentially producing one sequence at a time. For each input $x_i$ and at each output position $j$, SSG-S monitors the generated set $A$ of sequences so far (each with a length $j-1$). For each sequence $A_k$ in $A$, SSG-S generates all possible tokens $\vec{Y}_j$ at position $j$ by repeatedly finding the most probable solution and penalizing it. In other words, the testing procedure is similar to  that of SSG, except for the explicit consideration of all the partial sequence $A_k$. Then, SSG-S appends each token in $\vec{Y}_j$ to the corresponding $A_k$, producing new sequences $A_k$ with length $j$. Note that the previously generated answers in $A_k$ are used as context in the overall generation process.
It can be achieved by feeding $A_k$ into the decoder as input for the next token, a procedure similar to ``teacher forcing" in training seq2seq. With this gradual expansion of the answer set $A$, SSG-S produces all the feasible sequences.   

\section{Deep Sequential Set Generation}
\begin{figure*}[h]
\begin{center}
\centerline{\includegraphics[width=1.43\columnwidth, scale=1]{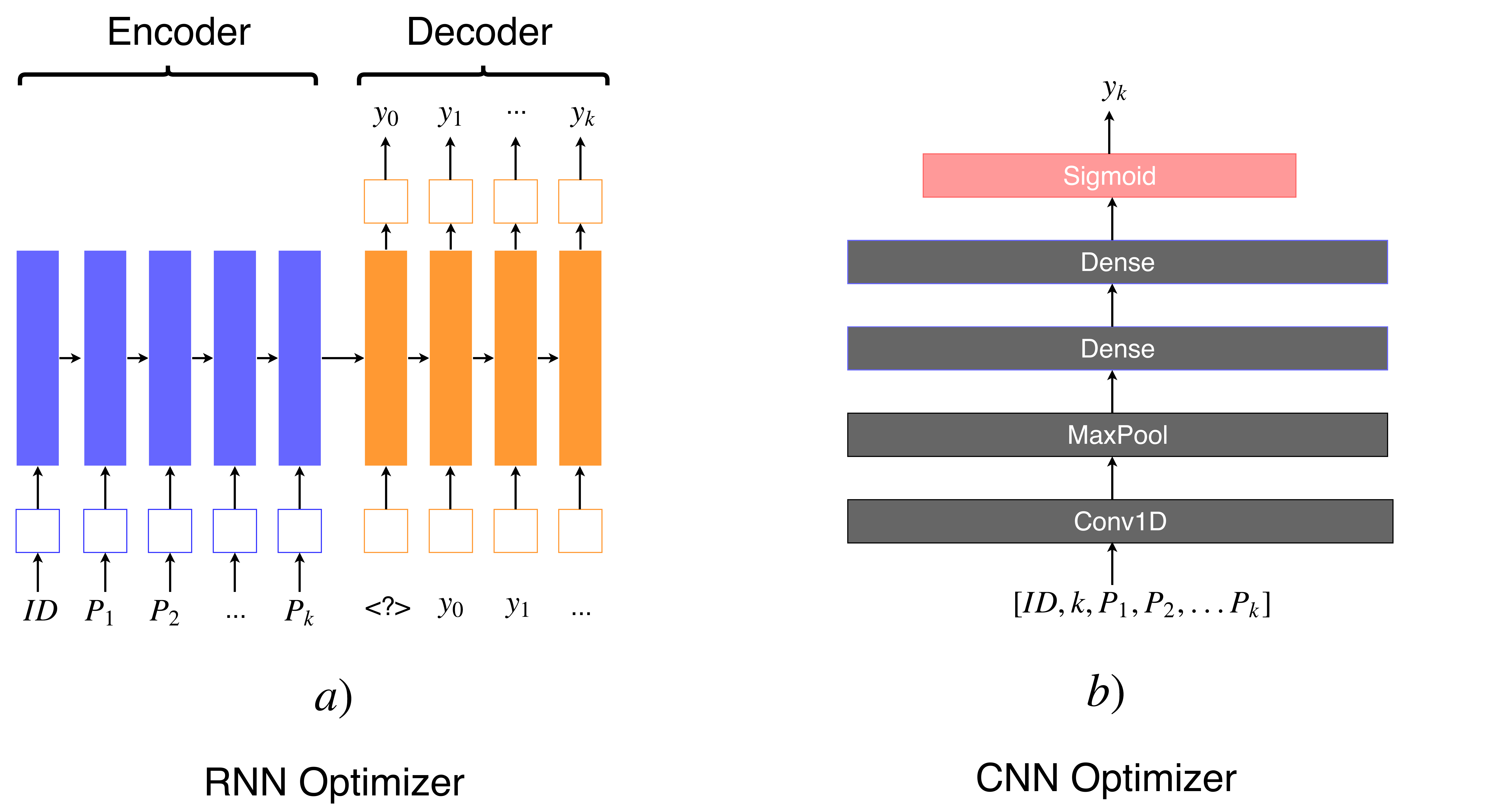}}
\caption{Architectures of RNN and CNN $\lambda$-Optimizer, as a part of SSG-S. Instead of learning $\lambda$'s directly, a neural-network-based classifier is used. Subfigure $(a)$ shows the RNN architecture, and $(b)$ shows the CNN architecture used in the experiments. }
\label{fig:nn}
\end{center}
\end{figure*}

While SSG-S handles short sequences quite well, in practice data can be unbalanced and have increasing complexity for long sequences and large vocabulary. The loss for different correct outputs in a set can hence substantially differ, depending on the label frequencies at each position of the sequence. This phenomenon could lead to a problem that one single $\lambda$, or even a fixed set of $\lambda$'s, cannot distinguish the positive and negative sets in different contexts.  To remedy this difficulty, we introduce a deep learning-based approach to distinguish the positive classes from the negative ones at each position $j$ in the sequence, replacing the learning of all $\lambda$'s as discussed in the preceding section. In essence, we use a neural network to classify positive and negative tokens in the sequence. Specifically, we still train a seq2seq model as discussed previously. However, now we feed the loss sequence in the final output layer into another neural network, which we call the $\lambda$-network. $\lambda$-network classifies  each possible label from the original network into either positive or negative class at that token value. During training, the $\lambda$-network is learned by taking the loss from the decoder logits  as inputs, and produces a binary label (indicating whether each label is a positive class) at position $j$.  We consider both recurrent neural networks (RNN) and convolutional neural networks (CNN) as the classifier. Their structures are shown in Figure~\ref{fig:nn}.

For the RNN $\lambda$-optimizer, we use another seq2seq model. We feed the decoder logits and the position ID of the desired target sequence as an input to the encoder part of the RNN, and then use the binary labels on each logit as training target for the decoder.  For the CNN $\lambda$-optimizer, we feed decoder logits and the position ID $j$ as well as the logit ID $k$, and use  one 1D-convolution and max pooling layers, multiple densely connected layers, and one sigmoid layer. The output of CNN is the binary label of $k$-th element of the logit.  Note that the $\lambda$-network only replaces the learning of $\lambda$ in Algorithm~\ref{alg:trains} and Equation~\eqref{eq:objs} of Algorithm~\ref{alg:tests}. The rest of the training and testing algorithms remain unchanged. We call the methods respectively SSG-RNN and SSG-CNN. Note that SSG-S along with SSG-RNN and SSG-CNN can both be used for the singleton sets, which can be considered as sequences of length 1.

\section{Experiments}
\label{sec:exp}

We conduct experiments to evaluate the proposed algorithms  on various applications, comparing against existing baselines if possible.

\subsection{Benchmark Dataset}

While it is not the intended application of the proposed sequential set generation algorithms, SSG can be applied to multi-label problems. We compare SSG with standard multi-label techniques on the YEAST and SCENE dataset, both of which are publicly available. YEAST is in the domain of biology. It contains over $2000$ data samples and has the feature size of $103$.  The unique label number is $14$, and the average cardinality is $4.2$.  The SCENE data has $2407$ samples, $294$ features, and $6$ unique labels.

We compare with the standard sigmoid network \cite{grodzicki2008improved}, where each possible label is considered as a binary classification problem. For fair comparison, we use the same base architecture for both the sigmoid network and deep SSG models, and take the sigmoid output as the input to $\lambda$-optimizer in SSG. Since the baseline consists of deep models, we only compare deep versions of SSG. We do a train-test split of $70-30$, and use the standard $F1$ score to measure the accuracy performance of different methods.  We then take the mean, $mF1$, as the accuracy score to compare the ground truth label set and the learned set. The higher the $mF1$ score, the better.

\begin{table}[h]
\centering
\caption{Mean $F_1$ Accuracy Result on Benchmark Dataset of Various Algorithms. SSG-CNN shows the best performance.}
\label{table:exp1}
\begin{tabular}{|l||l| l|l|l}
\cline{1-4}
  \rule{0pt}{4ex}      & Multi-Label  & SSG-RNN & SSG-CNN &  \\ \cline{1-4}
  \rule{0pt}{2ex} YEAST &  0.430 &     0.402    &    {\bf  0.658}   &  \\ 
 \cline{1-4}
 \rule{0pt}{2ex} SCENE & 0.455 &    0.378    & {\bf 0.605} \\ \cline{1-4}
\end{tabular}
\end{table}

\begin{table*}[!htbp]
\centering
\caption{Experiment Accuracy Results of Various Algorithms on Two Complex Reasoning Tasks involving Set Output and Set of Sequences. \newline Mean $F_1$ score (mF1) and mean Edit Distance (mED) are Used. }
\label{table:res}
\begin{tabular}{|l||l||l| l|l|l|l}
\cline{1-6}
  \rule{0pt}{4ex}       &Metric & Multi-Label & SSG-S & SSG-RNN & SSG-CNN &  \\ \cline{1-6}
\rule{0pt}{2ex}    Task 1 & $mF1$, the higher the better & 0.64 &  0.19   &    0.42     &    {\bf  0.70}   &  \\ 
 \cline{1-6}
\rule{0pt}{2ex}    Task 2 & $mED$, the lower the better & N/A & 8.10     &  3.75      &   {\bf  2.00}     &  \\ \cline{1-6}
\end{tabular}
\end{table*}

As one can see from Table~\ref{table:exp1}, SSG  substantially outperforms the simple sigmoid network for multi-label classification. Although one might use different or more complex architectures than the sigmoid network, we believe the relative improvement would be consistent (which supported in the following more complex tasks). 

\subsection{Synthetic Datasets}
We conduct two experiments to compare the proposed methods: a number problem that predicts sets, and another problem that predicts a set of sequences. We first describe each problem, with the aim of tackling complex reasoning tasks that traditional machine learning methods cannot handle.  

\noindent\textbf{Task 1: Predicting Sets}. 
In this task, the input is a positive integer read as a string of digits. Let the leading digit be $m$. The output is the set of $m$ leading digits of the input string, with duplicates counted only once. For example, if $X=33874$, then $Y=\{3,8\}$. We call this Task-1. We again use $mF1$ as the accuracy score to compare the ground truth label set and the learned set.

\noindent\textbf{Task 2: Predicting Set of Sequences}. In the second task, the input is a digit string of length 20. Let the string be evenly split into two halves. The first 10 digits are grouped into five pairs: $(s_1,e_1)$, \ldots, $(s_5,e_5)$; and the last 10 digits constitute a string $a$. The output set consists of (at most) 5 subsequences of $a$: $a[s_1,e_1)$, \ldots, $a[s_5,e_5)$. Whenever $s_i\ge e_i$ for some $i$, the substring is empty and hence it does not count as an element of the output set. Similar to the first data set, duplicate strings are removed. For example, if $X=00490000349172105519$, then $Y=\{2, 10551\}$. The elements of $Y$ are substrings $a[4,9)$ and $a[3,4)$, where $a=9172105519$. Note that 0-based indexing is used here. Treated as a multi-label classification problem, the number of classes is $10^{10}$, which is impossible to handle. We call this Task-2. We use mean edit distance, $mED$, as the accuracy score to compare the ground truth set of sequences and the learned set of sequences. For ground truth set and learned set, we compute $ED$ distance between every pair of sequences and divided by the total number of pairs.  The lower the score $mED$, the better. 

\noindent \textbf{System Architecture}: Since both tasks have sequence inputs, we use an encoder-decoder architecture~\cite{sutskever2014sequence}. We use a one layer LSTM  with $60$ encoder hidden units and $120$ decoder hidden units. An embedding layer of size $60$ is  used for appropriate discrete inputs and outputs. We use Adam optimizer~\cite{kingma2014adam}  with a batch size of $15$, and cross entropy as loss function. We generate 1000 samples and randomly split $70\%$ as training and the rest as testing.

We compare three methods SSG-S, SSG-RNN, and SSG-CNN with the baseline multi-label sigmoid network for these two tasks. Table~\ref{table:res} shows the results. In both tasks, we can see that SSG-CNN is the best method, outperforming the second best SSG-RNN by a large margin (28\% $mF1$ and 1.75 $mED$). Moreover,  the neural-network-based SSG-CNN and -RNN outperform SSG-S, showing that it is very important to consider the complexity of reasoning tasks. Note that we did not tune or search for the best hyper-parameters and it is reasonable to assume that these performance figures can  be  further improved. SSG-CNN also outperforms the multi-label method on Task $1$, and the multi-label method is not applicable to Task $2$ due to the extreme modeling complexity. 

\section{Conclusion}
We proposed a general framework, SSG, along with three variants, designed to solve set-valued output problems. We developed a sequential generation approach that can efficiently learn set relationships from data, as demonstrated on benchmark and reasoning tasks.  Experiments show that the sequential generation procedure can improve performance on traditional multi-label tasks and can handle more complex sets such as set of sequences, where traditional methods are not readily applicable. 

Further work will include theoretical analysis on the relationships between the set size and the learning performance, investigation on better training methods for SSG, and testing on a wider variety of set components, including sets of sets. We believe set-valued outputs have many applications such as theorem proving in AI and are foundational for systems that perform reasoning in particular, making their general treatment an important research direction to address. 


\section{Acknowledgments}
We thank for colleagues at AISR for helpful discussion and anonymous reviewers for insightful comments.

\bibliographystyle{aaai}
\bibliography{ijcai18}

\end{document}